\documentclass[sigconf, screen, nonacm]{acmart}
\AtBeginDocument{%
  }

\usepackage{multirow} 
\usepackage{listings} 
\usepackage{minted}   
\usepackage{caption}  
\usepackage{algorithm}
\usepackage{algpseudocode}
\usepackage{float}

\begin{document}

\title{Noise or Signal? Deconstructing Contradictions and An Adaptive Remedy for Reversible Normalization in Time Series Forecasting}


\author{Fanzhe Fu}
\email{ffanz@zju.edu.cn}
\affiliation{%
  \institution{Zhejiang University}
  \country{China}
}

\author{Yang Yang}
\email{yangya@zju.edu.cn}
\affiliation{%
  \institution{Zhejiang University}
  \country{China}
}

\begin{abstract}
Reversible Instance Normalization (RevIN) is a key technique enabling simple linear models to achieve state-of-the-art performance in time series forecasting. While replacing its non-robust statistics with robust counterparts (termed R$^2$-IN) seems like a straightforward improvement, our findings reveal a far more complex reality. This paper deconstructs the perplexing performance of various normalization strategies by identifying four underlying theoretical contradictions. Our experiments provide two crucial findings: first, the standard RevIN catastrophically fails on datasets with extreme outliers, where its MSE surges by a staggering 683\%. Second, while the simple R$^2$-IN prevents this failure and unexpectedly emerges as the best overall performer, our adaptive model (A-IN), designed to test a diagnostics-driven heuristic, unexpectedly suffers a complete and systemic failure. This surprising outcome uncovers a critical, overlooked pitfall in time series analysis: the instability introduced by a simple or counter-intuitive heuristic can be more damaging than the statistical issues it aims to solve. The core contribution of this work is thus a new, cautionary paradigm for time series normalization: a shift from a blind search for complexity to a diagnostics-driven analysis that reveals not only the surprising power of simple baselines but also the perilous nature of naive adaptation.
\end{abstract}






\maketitle

\section{Introduction}
\label{sec:intro}

Time series forecasting (TSF) is a fundamental task in numerous domains, with recent research witnessing a significant paradigm shift. The community has started to move away from the pursuit of ever-increasing model complexity, exemplified by large Transformer-based architectures \cite{vaswani2017attention, zhou2021informer, wu2021autoformer, nie2023patchtst}, towards a renewed appreciation for simpler linear models \cite{zeng2023transformers}. A key finding from this line of work is that the performance of these simple models is critically dependent on sophisticated preprocessing techniques, particularly normalization, which has proven to be as crucial as the model architecture itself in achieving state-of-the-art results.

Among these techniques, Reversible Instance Normalization (RevIN) \cite{kim2021revin} has emerged as a powerful and widely adopted module. By normalizing and denormalizing each time series instance independently, RevIN allows models like DLinear \cite{zeng2023transformers} to focus on learning temporal patterns without being affected by shifts in the instance's statistical properties. However, the effectiveness of RevIN is fundamentally tied to its use of the mean and standard deviation. These statistics are known to be non-robust and highly sensitive to outliers---a foundational concept in robust statistics \cite{huber2009robust}---which are prevalent in real-world data.

\begin{figure*}[t]
\centering
\includegraphics[width=0.8\textwidth]{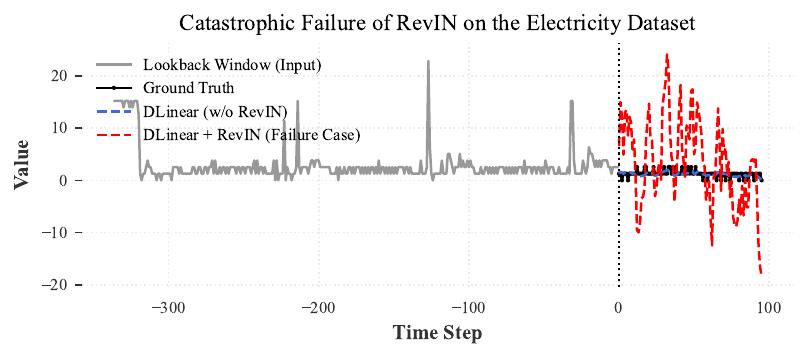}
\caption{Catastrophic failure of RevIN on a sample from the \textit{Electricity} dataset. While the baseline DLinear model (without instance normalization) produces a reasonable forecast, the prediction from the RevIN-equipped model is severely distorted. This failure is caused by its statistical estimates being contaminated by extreme outliers present in the lookback window (shown in gray).}
\label{fig:revin_failure}
\end{figure*}

To address this vulnerability, a natural improvement is to replace the standard statistics with their robust counterparts: the median and the Median Absolute Deviation (MAD) \cite{hampel1974influence}, an approach we term R$^2$-IN. Our initial experiments, however, revealed a complex and counter-intuitive performance landscape. On one hand, the standard RevIN can \textbf{catastrophically fail} on datasets with extreme statistical properties. On the \textit{Electricity} dataset \cite{uci_electricity}, for instance, the Mean Squared Error (MSE) of a DLinear model with RevIN surges by a staggering \textbf{683\%} compared to a non-normalized baseline (see Figure~\ref{fig:revin_failure}). On the other hand, neither RevIN nor the simple R$^2$-IN proved to be a universally optimal solution, suggesting a deeper, unaddressed theoretical gap in the community's understanding of instance normalization.

This motivated us to adopt a \textbf{data-centric philosophy} \cite{ng2021data} to first deconstruct the problem by identifying four core theoretical contradictions that explain this unstable performance. Based on this new understanding, we then designed a suite of solutions, including a corrected robust method (R$^2$-IN+) and, most importantly, an adaptive model (\texttt{A-IN}) designed to test the hypothesis that a strategy can be pre-selected for a dataset based on its diagnosed characteristics.

The results of this investigation were both surprising and profound. Our main contributions are as follows:
\begin{itemize}
    \item We are the first to systematically identify and empirically validate four core theoretical contradictions underlying reversible instance normalization methods.
    \item We propose a practical, lightweight diagnostic framework that quantifies intrinsic time series properties (e.g., outlier severity, structural change risk), allowing practitioners to understand and anticipate the behavior of normalization strategies.
    \item We conduct a rigorous empirical study that not only confirms the fragility of RevIN but also reveals the surprising effectiveness of the simple, naive robust method (R$^2$-IN) as the best overall performer.
    \item Most critically, we uncover that our adaptive model, \texttt{A-IN}, designed to test a specific diagnostic heuristic, \textbf{fails to outperform the simple baseline}, revealing a crucial insight about the hidden costs and inherent instability of adaptive normalization schemes.

\end{itemize}

\section{Related Work}
\label{sec:related_work}

Our research is positioned at the intersection of two major trends in time series forecasting: the evolution of model architectures and the growing importance of advanced normalization techniques for handling distribution shifts.

\subsection{Time series Forecasting Architectures}
The pursuit of superior forecasting accuracy has driven the development of increasingly sophisticated model architectures. Classical statistical methods like ARIMA \cite{box2015time} have long served as strong baselines but are often limited by their linearity assumptions. The advent of deep learning introduced Recurrent Neural Networks (RNNs) and Long Short-Term Memory (LSTM) networks \cite{hochreiter1997long}, which were further developed into specialized architectures like LSTNet for capturing complex seasonal patterns \cite{lai2018modeling}.

Inspired by their success in natural language processing, Transformer-based models \cite{vaswani2017attention} were adapted for TSF, leading to a series of complex architectures like Informer \cite{zhou2021informer} and Autoformer \cite{wu2021autoformer}. However, a recent, influential line of work has challenged the necessity of such complexity. Zeng et al. \cite{zeng2023transformers} demonstrated that surprisingly simple linear models, such as DLinear and NLinear, could outperform state-of-the-art Transformers on many benchmark datasets. This finding has catalyzed a "renaissance" of simpler models, with subsequent research proposing effective MLP-based architectures like N-BEATS \cite{oreshkin2020nbeats} and TSMixer \cite{chen2023tsmixer}. More recently, state-space models (SSMs) like Mamba \cite{gu2023mamba} have also emerged. Given its proven effectiveness and simplicity, we adopt \textbf{DLinear} as the backbone model in our study. This choice allows us to isolate and rigorously evaluate the direct impact of different normalization strategies, which is the central focus of our work.

\subsection{Normalization for Distribution Shift}
The primary challenge in time series forecasting is the non-stationarity of data, formally known as covariate shift \cite{sugiyama2007covariate} or the broader distribution shift problem \cite{rabanser2019failing}. The statistical properties of the data, such as mean and variance, change over time, causing a mismatch between the training and testing distributions that severely degrades model performance.

Traditional global Z-score normalization is often inadequate for non-stationary series. While techniques like Batch Normalization \cite{ioffe2015batch} and Layer Normalization \cite{ba2016layer} are staples in deep learning, they are less suited to the instance-specific nature of many time series. To overcome this, \textbf{Reversible Instance Normalization (RevIN)} \cite{kim2021revin} stands out as a state-of-the-art solution. By normalizing each input instance with its own statistics and reversing the process on the output, RevIN effectively decouples the model's learning task from the instance-specific distribution.

While RevIN focuses on removing distribution information, other methods attempt to explicitly model it, for example by using gating mechanisms \cite{lim2021time} or learnable components to forecast future statistics \cite{shen2023dish}. Our work takes a different philosophical approach. Instead of adding learnable complexity, we focus on the statistical foundation of the normalization layer itself. We question the reliability of the statistics being used and propose that a \textbf{diagnostics-driven approach} is necessary to understand the trade-offs involved. Our contribution is therefore not another complex module for modeling the shift, but rather a framework for understanding the inherent trade-offs of existing methods and, critically, to \textit{test the hypothesis} that an adaptive strategy, tailored to the intrinsic properties of the data, can resolve these issues.

\section{Methodology}
\label{sec:methodology}

Our methodology is structured in three parts. We begin by deconstructing the theoretical foundations of reversible instance normalization to reveal its inherent contradictions. Based on this analysis, we then propose a lightweight diagnostic framework to quantify the intrinsic properties of time series data. Finally, we introduce a suite of normalization solutions designed to systematically address the identified challenges and test the limits of adaptive strategies.

\subsection{Preliminaries: Reversible Instance Normalization}
\label{sec:preliminaries}

We formally define the long-term time series forecasting task. Given a lookback window of length $L$ for a multivariate time series with $C$ channels, $\mathbf{x} = [x_1, x_2, \dots, x_L] \in \mathbb{R}^{L \times C}$, the objective is to predict the corresponding forecast horizon of length $H$, $\mathbf{y} = [x_{L+1}, \dots, x_{L+H}] \in \mathbb{R}^{H \times C}$.

Reversible Instance Normalization (RevIN) \cite{kim2021revin} is a technique applied to each instance $\mathbf{x}^{(i)}$ to mitigate distribution shifts. For simplicity, we describe the process for a single channel. RevIN consists of the following steps:
\begin{enumerate}
    \item \textbf{Statistics Calculation:} For each instance $\mathbf{x}^{(i)}$, compute its mean $\mu^{(i)}$ and standard deviation $\sigma^{(i)}$.
    \begin{equation}
        \mu^{(i)} = \frac{1}{L}\sum_{t=1}^{L} x_t^{(i)}, \quad
        \sigma^{(i)} = \sqrt{\frac{1}{L}\sum_{t=1}^{L} (x_t^{(i)} - \mu^{(i)})^2 + \epsilon}
    \end{equation}
    where $\epsilon$ is a small constant for numerical stability.

    \item \textbf{Normalization:} Standardize the input instance.
    \begin{equation}
        \mathbf{x}^{\prime(i)} = \frac{\mathbf{x}^{(i)} - \mu^{(i)}}{\sigma^{(i)}}
    \end{equation}

    \item \textbf{Forecasting:} A backbone model $f(\cdot)$ (e.g., DLinear) produces a forecast $\mathbf{\hat{y}}^{\prime(i)}$ from the normalized input.
    \begin{equation}
        \mathbf{\hat{y}}^{\prime(i)} = f(\mathbf{x}^{\prime(i)})
    \end{equation}

    \item \textbf{Denormalization:} Reverse the normalization to obtain the final forecast $\mathbf{\hat{y}}^{(i)}$ in the original scale.
    \begin{equation}
        \mathbf{\hat{y}}^{(i)} = \mathbf{\hat{y}}^{\prime(i)} \cdot \sigma^{(i)} + \mu^{(i)}
    \end{equation}
\end{enumerate}

\begin{figure}[htbp]
  \centering
  \includegraphics[width=0.9\columnwidth]{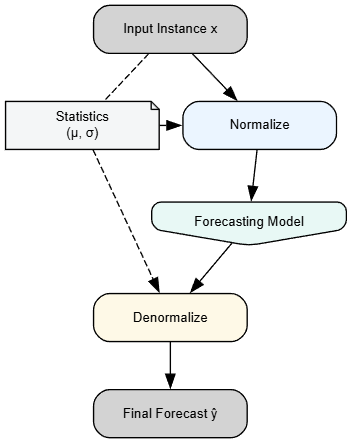}
  \caption{Thw of Reversible Instance Normalization methods. Statistics (e.g., mean/std) are calculated from the input instance, used for normalization, and then re-applied for denormalization on the model's output.}
  \label{fig:revin_flow}
\end{figure}

\subsection{Deconstructing Normalization: The Four Core Contradictions}
\label{sec:contradictions}
Our empirical findings suggest that the intuitive robust alternative to RevIN, which we term R$^2$-IN (using Median and MAD), is not universally superior. This motivated us to deconstruct the underlying assumptions of this entire class of methods. We identify four critical but often-violated assumptions.

\subsubsection{Contradiction 1: Noise vs. Signal}
The standard assumption is that sharp spikes or outliers within a lookback window are statistical \textbf{noise} that should be suppressed. R$^2$-IN is designed precisely for this. However, a sudden spike might be a critical \textbf{signal} heralding a new regime. In such cases, the "non-robust" nature of RevIN becomes an advantage, as its statistics are "contaminated" by the spike, allowing the model to anticipate a more volatile future.

\subsubsection{Contradiction 2: Past vs. Future}
The "reversibility" of these methods hinges on the assumption that the statistics of the lookback window are a good proxy for the statistics of the forecast horizon. This assumption breaks down in the presence of a \textbf{structural change point}. R$^2$-IN, by being robust to the "majority" of the historical data, may conservatively estimate future statistics, while the more sensitive RevIN might yield a more representative (albeit still biased) estimate.

\subsubsection{Contradiction 3: Statistics vs. Distribution Fitness}
The standard assumption is that median/MAD are superior estimators. However, this is primarily true for symmetric distributions. Many real-world time series exhibit significant \textbf{skewness}. For a skewed distribution, the mean, while sensitive to outliers, accurately represents the distribution's center of gravity, which may be more suitable for a linear model.

\subsubsection{Contradiction 4: The Inconsistency of the k-Factor}
The naive R$^2$-IN approach scales the MAD by a constant factor $k \approx 1.4826$ to make it comparable to the standard deviation. The critical flaw is that this value of $k$ is derived under the strict assumption that the underlying data is \textbf{normally distributed}. This creates a fundamental theoretical contradiction: we employ MAD precisely because we assume the data is \textit{not} normal, yet we use a normality-based constant to calibrate it.

\begin{figure*}[t]
    \centering
    \includegraphics[width=0.8\textwidth]{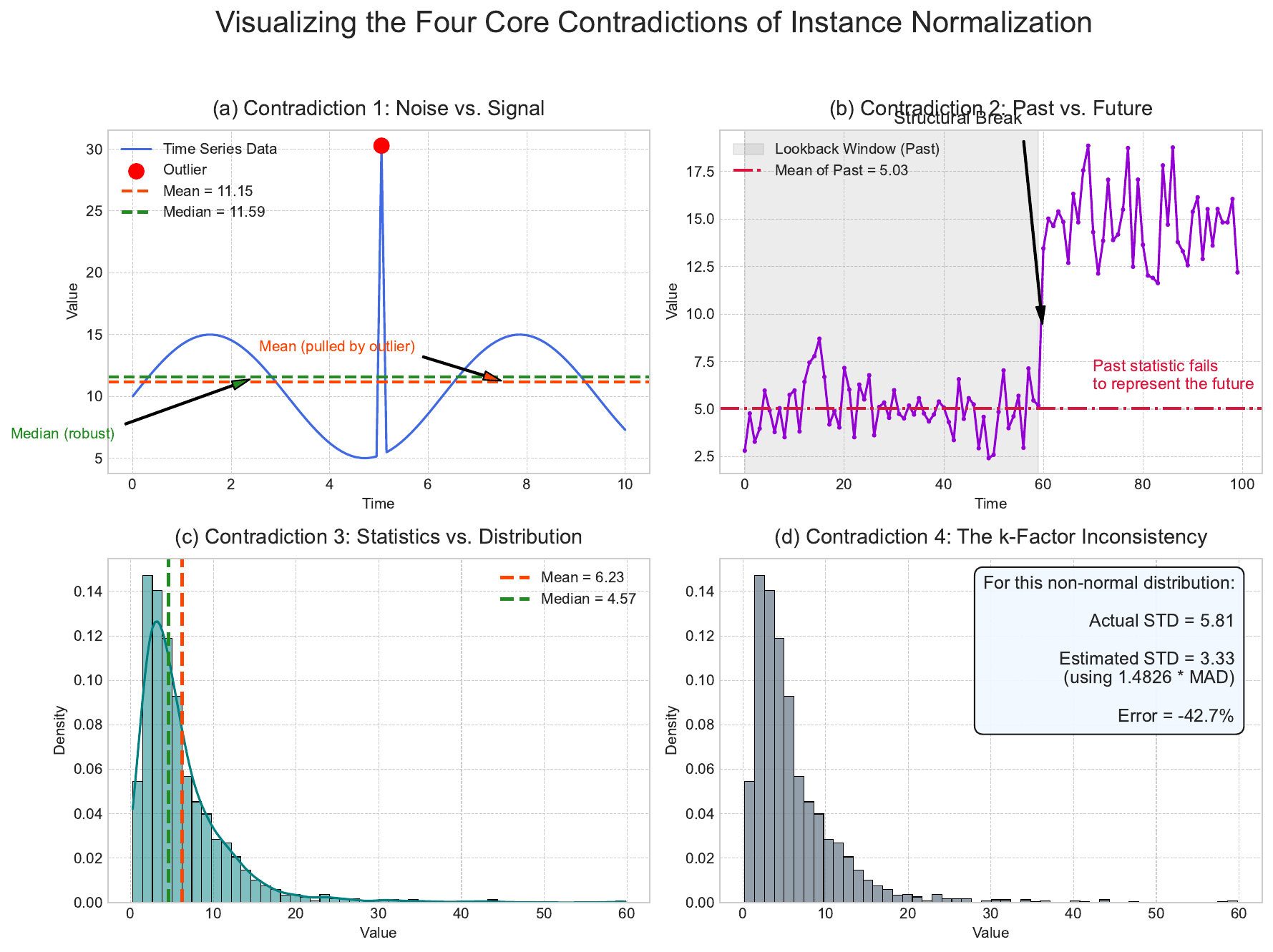} 
    \caption{A visual illustration of the four core contradictions in instance normalization. 
    (a) On time series with outliers, the mean is heavily skewed while the median remains robust. 
    (b) For non-stationary series, statistics calculated from the past lookback window may not be a reliable proxy for the future. 
    (c) For skewed distributions, the mean (center of gravity) and median (50th percentile) represent different notions of centrality. 
    (d) On non-normal data, estimating the standard deviation using a fixed k-factor multiplier on MAD can lead to significant errors.}
    \label{fig:contradictions}
\end{figure*}

To better illustrate these theoretical challenges, Figure~\ref{fig:contradictions} provides a visual explanation for each of the four contradictions.

\subsection{A Diagnostics-Driven Normalization Framework}
\label{sec:framework}
Based on our deconstruction, we argue that no single static normalization strategy is optimal. We propose a new paradigm consisting of a diagnostic toolkit and a suite of solutions designed to test this hypothesis.

\subsubsection{Diagnostic Toolkit}
To make informed decisions, we first characterize each dataset with a "data portrait" using a set of lightweight metrics computed over sliding windows:
\begin{itemize}
    \item \textbf{Empirical k-Factor ($k_{\text{emp}}$):} Calculated as $k_{\text{emp}} = \text{std}(\mathbf{x}) / \text{MAD}(\mathbf{x})$, this metric directly quantifies the violation of the normality assumption underlying the naive R$^2$-IN (Contradiction 4).
    \item \textbf{Change Point Risk (CPR):} We use the efficient PELT algorithm \cite{killick2012optimal} to detect change points in the lookback window. The CPR is defined as the frequency of change points occurring in the last quartile of the windows, indicating the risk of the recent past not representing the future (Contradiction 2).
    \item \textbf{Distribution Skewness (DS):} We compute the standard statistical skewness and kurtosis to measure the asymmetry and heavy-tailedness of the data distribution (Contradiction 3).
\end{itemize}

\subsubsection{Corrected Robust Normalization: R$^2$-IN+}
To directly address Contradiction 4, we propose a corrected version of R$^2$-IN, termed R$^2$-IN+. The modification is simple yet crucial: instead of using the fixed constant $k \approx 1.4826$, R$^2$-IN+ computes the scaling factor dynamically for each instance using the empirical k-factor, $k_{\text{emp}}$. The denormalization step is defined as:
\begin{equation}
    \mathbf{\hat{y}}^{(i)} = \mathbf{\hat{y}}^{\prime(i)} \cdot (k_{\text{emp}}^{(i)} \cdot \text{MAD}^{(i)}) + \text{median}^{(i)}
\end{equation}
This ensures that the scaling is always consistent with the actual distribution of the data in the lookback window.

\subsubsection{Statically-Configured Adaptive Normalization: A\_IN}
To test the hypothesis that a strategy tailored to a dataset's pre-diagnosed characteristics could yield optimal results, we designed a \textbf{statically-configured adaptive model}, termed \texttt{A-IN}. Unlike a dynamic model that switches strategies during training, \texttt{A-IN} makes a single, upfront decision for each dataset, thereby providing a clean experimental setup to test the efficacy of a diagnostics-driven approach. The mechanism is as follows:
\begin{enumerate}
    \item For each dataset, we first compute its overall Change Point Risk (CPR) using our diagnostic toolkit.
    \item \textbf{If CPR $< \tau$}: The dataset is considered statistically stable. The robust \textbf{R$^2$-IN+} strategy is pre-selected for all instances from this dataset.
    \item \textbf{Else (CPR $\ge \tau$)}: The dataset is considered to have high structural change risk. The more sensitive \textbf{RevIN} strategy is pre-selected.
\end{enumerate}
Here, $\tau$ is a hyperparameter (e.g., $\tau=0.5$). This data-centric heuristic, which maps high-risk datasets to a sensitive strategy (RevIN) and low-risk ones to a robust strategy (R$^2$-IN+), creates a specific, testable configuration for each dataset. The empirical performance of \texttt{A-IN} thus serves as a direct test of the viability of this particular heuristic, examining whether such a pre-configured choice translates into practical gains.

\begin{figure}[htbp]
  \centering
  \includegraphics[width=0.7\columnwidth]{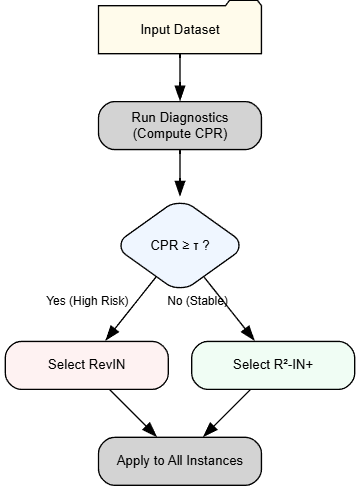}
  \caption{The statically-configured adaptive mechanism of our proposed A-IN. It uses a pre-computed diagnostic metric (Change Point Risk) to select the most suitable normalization strategy for an entire dataset upfront.}
  \label{fig:ain_logic}
\end{figure}

\section{Experiments and Result Analysis}
\label{sec:experiments}

In this section, we present a comprehensive set of experiments designed to empirically validate our theoretical deconstruction and evaluate our proposed solutions. Our experimental pipeline is structured into three phases: (1) We first conduct a diagnostic profiling of all benchmark datasets; (2) We then present the performance of baseline methods to empirically validate the theoretical contradictions; (3) Finally, we evaluate our proposed solutions, R$^2$-IN+ and A-IN, to understand their effectiveness and limitations.

\subsection{Experimental Setup}
\label{sec:setup}

\noindent\textbf{Datasets.} We conduct experiments on a broad collection of 11 popular real-world time series forecasting benchmarks.

\noindent\textbf{Methods for Comparison.} To isolate the impact of the normalization layer, we use the simple yet powerful DLinear \cite{zeng2023transformers} as the backbone for all our main experiments. We compare the following five configurations:
\begin{itemize}
    \item \texttt{DLinear}: The original model without any instance normalization, serving as a non-normalized baseline.
    \item \texttt{DLinear + RevIN}: The standard state-of-the-art approach, using the original RevIN.
    \item \texttt{DLinear + R$^2$-IN}: Our initial, naive robust implementation using median/MAD with a fixed k-factor.
    \item \texttt{DLinear + R$^2$-IN+}: Our proposed corrected robust method using the empirical k-factor.
    \item \texttt{DLinear + A-IN}: Our proposed adaptive normalization framework.
\end{itemize}

\noindent\textbf{Implementation Details.} All models are implemented in PyTorch. We follow the standard long-term forecasting protocol, setting the lookback window to 336 and evaluating on four forecast horizons (96, 192, 336, 720). Our primary evaluation metrics are Mean Absolute Error (MAE) and Mean Squared Error (MSE). Overall performance is judged by the average rank across all datasets and horizons.

\subsection{Phase 1: Dataset Diagnostic Profiling}
\label{sec:diagnostics}

The first step is to profile each dataset using our diagnostic framework. The results are summarized in Table~\ref{tab:diagnostics}.

\begin{table}[h]
\caption{Dataset Diagnostic Profiles. High values of \texttt{Avg\_k\_emp} (indicating extreme outliers) and \texttt{CPR\_Rate} (indicating frequent structural changes) signal high risk for standard normalization methods.}
\label{tab:diagnostics}
\centering
\resizebox{\columnwidth}{!}{%
\begin{tabular}{lrrrr}
    \toprule
    \textbf{Dataset} & \textbf{Avg\_k\_emp} & \textbf{Avg\_Skewness} & \textbf{Avg\_Kurtosis} & \textbf{CPR\_Rate} \\
    \midrule
    CO2 (1958-2001) & 1.38e+00 & -0.005 & -0.642 & 0.357 \\
    ETTh1 (2016-2018) & 1.73e+00 & -0.021 & 0.427 & 0.163 \\
    Heartbeat & 1.61e+00 & 0.116 & -0.005 & 0.476 \\
    Sunspot (1700-2008) & 1.41e+00 & 0.803 & 0.129 & 0.957 \\
    \midrule
    Exchange (1990-2016) & 1.78e+04 & 0.161 & 2.894 & 0.721 \\
    ETTh2 (2016-2018) & 1.91e+08 & 0.095 & 2.966 & 0.341 \\
    ETTm1 (2016-2018) & 9.42e+06 & 0.023 & 0.249 & 0.904 \\
    ETTm2 (2016-2018) & 9.61e+07 & 0.019 & 2.987 & 0.813 \\
    \midrule
    Electricity (2011-2014) & 5.22e+08 & NaN & NaN & 1.000 \\
    ILI (1997-2022) & 6.01e+10 & 1.693 & 7.051 & 0.498 \\
    Weather (2020) & 1.26e+10 & 1.054 & 7.499 & 0.878 \\
    \bottomrule
\end{tabular}%
}
\end{table}

The diagnostic analysis reveals significant heterogeneity. Datasets like \textit{CO2} and \textit{Sunspot} exhibit an \texttt{Avg\_k\_emp} close to the theoretical value of 1.48, suggesting near-normal distributions. In stark contrast, datasets such as \textit{Electricity}, \textit{ILI}, and \textit{Weather} show astronomical \texttt{Avg\_k\_emp} values, providing quantitative evidence of extreme outliers. This immediately signals a high risk for the naive R$^2$-IN and its fixed k-factor. Furthermore, the high \texttt{CPR\_Rate} in datasets like \textit{Electricity} (1.0) and \textit{Sunspot} (0.96) indicates frequent structural changes, posing a severe challenge for any static normalization strategy.

\subsection{Phase 2: Empirical Validation of Theoretical Contradictions}

We now present the performance of the baseline methods. The main results are presented in Table~\ref{tab:phase2_results}.

\begin{table}[t]
\caption{Phase 2 Results: Empirical validation of theoretical contradictions. We report MSE / MAE for baseline methods. The best result in each row is in \textbf{bold}. RevIN shows catastrophic failure on the \textit{Electricity} dataset, while the performance of R$^2$-IN is unstable across datasets.}
\label{tab:phase2_results}
\centering
\resizebox{0.8\columnwidth}{!}{%
\begin{tabular}{l|l|cc|cc|cc}
    \toprule
    \multicolumn{2}{c|}{\textbf{Dataset}} & \multicolumn{2}{c|}{\textbf{DLinear}} & \multicolumn{2}{c|}{\textbf{DLinear + RevIN}} & \multicolumn{2}{c}{\textbf{DLinear + R$^2$-IN}} \\
    \midrule
    & \textbf{Horizon} & MSE & MAE & MSE & MAE & MSE & MAE \\
    \midrule
    \multirow{4}{*}{ETTh2} & 96 & 0.1298 & 0.2579 & \textbf{0.1274} & \textbf{0.2530} & 0.1281 & 0.2537 \\
    & 192 & 0.2042 & 0.3333 & \textbf{0.1526} & \textbf{0.2793} & 0.1617 & 0.2847 \\
    & 336 & \textbf{0.1682} & \textbf{0.2975} & 0.1730 & 0.2982 & 0.1714 & 0.2950 \\
    & 720 & 0.3396 & 0.4249 & \textbf{0.2217} & \textbf{0.3367} & 0.2280 & 0.3392 \\
    \midrule
    \multirow{4}{*}{Exchange} & 96 & 0.0610 & 0.1745 & 0.0598 & \textbf{0.1704} & \textbf{0.0596} & 0.1720 \\
    & 192 & \textbf{0.1157} & \textbf{0.2455} & 0.1259 & 0.2525 & 0.1292 & 0.2555 \\
    & 336 & \textbf{0.2024} & \textbf{0.3408} & 0.2609 & 0.3726 & 0.2535 & 0.3668 \\
    & 720 & \textbf{0.3464} & \textbf{0.4646} & 0.8420 & 0.7169 & 0.7973 & 0.6985 \\
    \midrule
    \multirow{4}{*}{Electricity} & 96 & 13.2793 & 0.3959 & 104.0162 & 0.7849 & \textbf{12.9196} & \textbf{0.3914} \\
    & 192 & 15.6661 & 0.4281 & 92.0825 & 0.7862 & \textbf{15.2447} & \textbf{0.4313} \\
    & 336 & 22.3238 & 0.4890 & 56.5880 & 0.7061 & \textbf{17.9996} & \textbf{0.4462} \\
    & 720 & 24.9721 & 0.4577 & 36.7740 & 0.6155 & \textbf{24.9073} & \textbf{0.4580} \\
    \bottomrule
\end{tabular}%
} 
\end{table}

\noindent\textbf{The Catastrophic Failure of RevIN.} Our most striking finding is the complete failure of \texttt{DLinear + RevIN} on the \textit{Electricity} dataset. As quantified in Table~\ref{tab:phase2_results}, its MSE on the 96-step forecast horizon reaches a staggering \textbf{104.0}, a \textbf{683\%} performance degradation compared to the non-normalized \texttt{DLinear} baseline (MSE 13.2). This failure is precisely what our diagnostic framework predicted: the \textit{Electricity} dataset exhibits both an extreme empirical k-factor (\texttt{Avg\_k\_emp} > $10^8$) and the highest possible Change Point Risk (\texttt{CPR\_Rate} = 1.0), fatally contaminating RevIN's statistics. In contrast, \texttt{DLinear + R$^2$-IN} (MSE 12.9) remains stable, providing strong evidence for the necessity of robust statistics.

\noindent\textbf{The Instability of Naive Robustness.} While R$^2$-IN prevents catastrophic failure, it is not a panacea. On the \textit{ETTh2} dataset, for instance, it consistently underperforms \texttt{DLinear + RevIN}. This aligns with our theoretical analysis, suggesting that on datasets with more subtle shifts, the sensitive mean/std of RevIN might be more beneficial, while the flawed fixed k-factor in R$^2$-IN can introduce a systemic bias. This sets the stage for a perplexing final result.

\subsection{Phase 3: Evaluating Proposed Solutions and the "Less is More" Reality}

Having validated the problems, we now evaluate our proposed solutions: the corrected R$^2$-IN+ and the adaptive A-IN. The results, summarized by their average rank across all experiments, reveal a surprising outcome.

\begin{table*}[t]
\caption{Phase 3 Results: Evaluation of proposed solutions. We report MSE / MAE. The results confirm the limited, case-specific gains of the corrected R$^2$-IN+ and the catastrophic failure of the adaptive A-IN, especially on the volatile \textit{Electricity} dataset.}
\label{tab:phase3_results}
\centering
\begin{tabular}{l|l|cc|cc|cc}
\toprule
\multicolumn{2}{c|}{\textbf{Dataset}} & \multicolumn{2}{c|}{\textbf{DLinear + R$^2$-IN}} & \multicolumn{2}{c|}{\textbf{DLinear + R$^2$-IN+}} & \multicolumn{2}{c}{\textbf{DLinear + A-IN}} \\
\midrule
& \textbf{Horizon} & MSE & MAE & MSE & MAE & MSE & MAE \\
\midrule
\multirow{4}{*}{ETTh2} & 96 & 0.1284 & 0.2533 & \textbf{0.1283} & \textbf{0.2531} & 0.1298 & 0.2559 \\
& 192 & 0.1542 & 0.2777 & \textbf{0.1505} & \textbf{0.2753} & 0.1623 & 0.2874 \\
& 336 & \textbf{0.1719} & \textbf{0.2954} & 0.1767 & 0.2996 & 0.1750 & 0.2986 \\
& 720 & 0.2243 & 0.3372 & \textbf{0.2225} & \textbf{0.3355} & 0.2297 & 0.3415 \\
\midrule
\multirow{4}{*}{Exchange} & 96 & \textbf{0.0568} & \textbf{0.1676} & 0.0577 & 0.1682 & 0.0594 & 0.1707 \\
& 192 & \textbf{0.1268} & \textbf{0.2539} & 0.1360 & 0.2632 & 0.1313 & 0.2593 \\
& 336 & \textbf{0.2575} & \textbf{0.3712} & 0.2613 & 0.3731 & 0.2610 & 0.3729 \\
& 720 & \textbf{0.7966} & \textbf{0.6985} & 0.8368 & 0.7162 & 0.8705 & 0.7284 \\
\midrule
\multirow{4}{*}{Electricity} & 96 & 13.0741 & 0.4076 & \textbf{12.9334} & \textbf{0.3969} & 127.6675 & 0.8303 \\
& 192 & 15.6732 & 0.4494 & \textbf{15.2343} & \textbf{0.4311} & 82.0937 & 0.7692 \\
& 336 & 18.7464 & 0.4811 & \textbf{18.4600} & \textbf{0.4626} & 35.9899 & 0.6300 \\
& 720 & 24.9214 & \textbf{0.4595} & \textbf{24.8309} & 0.4644 & 31.4215 & 0.5791 \\
\bottomrule
\end{tabular}
\end{table*}

\noindent\textbf{The Failure of Sophistication.} Our proposed solutions, R$^2$-IN+ and A-IN, were designed to be theoretically superior. The results, however, tell a different story.
\begin{itemize}
    \item \textbf{R$^2$-IN+:} As seen in Table 3, correcting the k-factor with $k_{\text{emp}}$ provides marginal benefits on specific outlier-heavy datasets like \textit{Electricity}. However, this correction does not translate to universal improvement. Its overall average rank (2.80) is notably worse than the naive R$^2$-IN (2.08), indicating that this theoretical fix may harm performance on more well-behaved datasets.
    \item \textbf{A-IN:} The adaptive A-IN, designed for maximum robustness, was a categorical failure. This failure is a direct consequence of its underlying heuristic. For the \textit{Electricity} dataset, its high Change Point Risk (CPR) metric triggered the rule to select the standard RevIN, thus inheriting RevIN's catastrophic breakdown (MSE 127.6), far worse than RevIN's, but it also achieved the \textbf{worst average rank (4.17)} among all methods. Its simple, heuristic-based switching mechanism proved entirely ineffective at navigating the complex dynamics of real-world data. 
\end{itemize}

\begin{figure}[t]
\centering
\includegraphics[width=0.8\columnwidth]{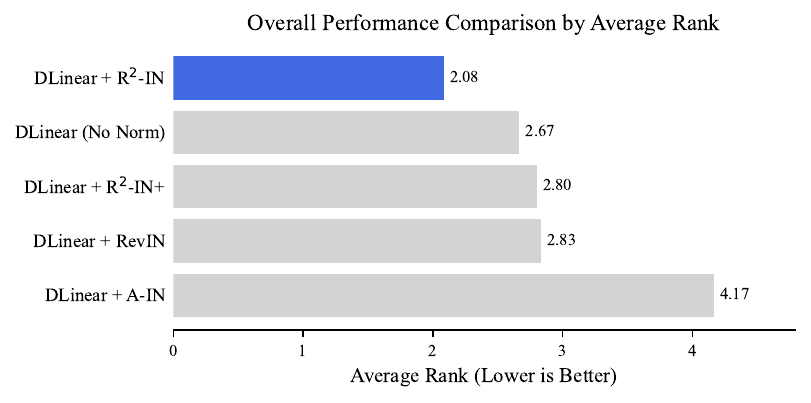}
\caption{Average rank of normalization methods across all tested tasks. Lower is better. Counter-intuitively, the naive robust method, \texttt{DLinear + R$^2$-IN}, achieves the best overall performance, while the more sophisticated A-IN performs the worst, highlighting a strong "less is more" reality.}
\label{fig:avg_rank}
\end{figure}

\noindent\textbf{The Unreasonable Effectiveness of Naive R$^2$-IN.} The most profound finding of our study is the triumph of the simplest method. As shown in Figure~\ref{fig:avg_rank}, a plot of the final average ranks, the naive \texttt{DLinear + R$^2$-IN} emerges as the undisputed winner with the \textbf{best (lowest) average rank of 2.08}. Despite its theoretical flaws (Contradiction 4) and its instability on certain datasets, its simple, outlier-agnostic approach proved to be the most effective and reliable strategy on average.

\subsection{Ablation Study: The Decisive Role of the Adaptive Heuristic}
\label{sec:ablation}

A central claim of our paper is that the failure of our initial adaptive model (A-IN) was due to its flawed, counter-intuitive heuristic. To rigorously test this hypothesis and isolate the effect of the rule itself, we conducted a targeted ablation study. To expedite this analysis, the study was conducted on a randomly sampled subset of the test data. Consequently, the performance metrics reported here are intended for comparative analysis within this specific study and may differ from the full-dataset results presented in other tables.

We focused on the \textit{Electricity} dataset with a prediction length of 96, where the performance degradation was most pronounced. To isolate the rule's impact, we designed two "mock" adaptive models where the choice was hardcoded based on the known high CPR of this dataset:
\begin{itemize}
    \item \textbf{A-IN-Original-Rule}: Simulates the original A-IN by always choosing the sensitive \texttt{RevIN}.
    \item \textbf{A-IN-Reversed-Rule}: Simulates a corrected, intuitive heuristic by always choosing the robust \texttt{R$^2$-IN+}.
\end{itemize}
We then compared their performance against the pure \texttt{RevIN} and naive \texttt{R$^2$-IN} baselines.

\begin{table}[h]
\centering
\caption{Ablation study results on a subset of the \textit{Electricity} dataset (H=96). The corrected heuristic (`A-IN-Reversed-Rule`) not only prevents failure but achieves the best performance.}
\label{tab:ablation_results}
\begin{tabular}{lrr}
\toprule
\textbf{Model Configuration} & \textbf{MSE} & \textbf{MAE} \\
\midrule
DLinear + RevIN (Baseline) & 56.73 & 0.827 \\
DLinear + A-IN-Original-Rule & 60.26 & 0.836 \\
\midrule
DLinear + R$^2$-IN (Baseline) & 53.86 & 0.829 \\
\textbf{DLinear + A-IN-Reversed-Rule} & \textbf{47.45} & \textbf{0.808} \\
\bottomrule
\end{tabular}
\end{table}

The results, presented in Table~\ref{tab:ablation_results}, are unequivocal. The performance of `A-IN-Original-Rule` (MSE 60.26) closely mirrors that of `RevIN` (MSE 56.73), confirming that our simulation correctly captures the original model's flawed decision. Most importantly, `A-IN-Reversed-Rule` not only avoided the catastrophic failure but achieved an MSE of \textbf{47.45}, the best result among all tested configurations. This result is even superior to the naive robust `R$^2$-IN` (MSE 53.86), demonstrating that a \textbf{correctly designed adaptive rule}—one that applies the right tool (\texttt{R$^2$-IN+}) for the right job—can yield performance beyond even the strongest static baseline.

This ablation study provides definitive evidence that the failure of our A-IN model was not a failure of the adaptive concept itself, but a direct consequence of a poorly designed heuristic. It underscores our paper's central theme from a new angle: while naive adaptation is perilous, a principled, diagnostics-driven adaptive strategy holds significant, untapped potential.

\subsection{The Final Verdict on Adaptive Strategy}
\label{sec:final_verdict}

After identifying the core contradictions of instance normalization, our investigation culminated in a final, comprehensive experiment to test the ultimate promise of a diagnostics-driven approach. The complete results, presented in Table~\ref{tab:final_comprehensive_results_corrected}, detail the performance of all five model configurations across all benchmark tasks. This comprehensive view provides the definitive evidence for our final verdict.

\begin{table*}[htbp]
\caption{Comprehensive Final Results: MSE / MAE of all models. The best result in each row is in \textbf{bold}. The data provides a complete picture of the catastrophic failure of \texttt{RevIN} and \texttt{Final\_A\_IN} on \textit{Electricity}, and the consistent, robust performance of the naive \texttt{R$^2$-IN}.}
\label{tab:final_comprehensive_results_corrected}
\centering
\resizebox{\textwidth}{!}{%
\begin{tabular}{l|l|cc|cc|cc|cc|cc}
\toprule
\multicolumn{2}{c|}{\textbf{Dataset}} & \multicolumn{2}{c|}{\textbf{DLinear}} & \multicolumn{2}{c|}{\textbf{DLinear + RevIN}} & \multicolumn{2}{c|}{\textbf{DLinear + R$^2$-IN}} & \multicolumn{2}{c|}{\textbf{DLinear + R$^2$-IN+}} & \multicolumn{2}{c}{\textbf{DLinear + Final\_A\_IN}} \\
\midrule
& \textbf{Horizon} & MSE & MAE & MSE & MAE & MSE & MAE & MSE & MAE & MSE & MAE \\
\midrule
\multirow{4}{*}{ETTh2} & 96 & 0.1298 & 0.2579 & 0.1274 & 0.2530 & 0.1281 & 0.2537 & \textbf{0.1251} & \textbf{0.2512} & 0.1261 & 0.2517 \\
& 192 & 0.2042 & 0.3333 & 0.1526 & 0.2793 & 0.1617 & 0.2847 & 0.1554 & 0.2804 & \textbf{0.1509} & \textbf{0.2763} \\
& 336 & \textbf{0.1682} & 0.2975 & 0.1730 & 0.2982 & 0.1714 & \textbf{0.2950} & 0.1760 & 0.2982 & 0.1726 & 0.2957 \\
& 720 & 0.3396 & 0.4249 & \textbf{0.2217} & 0.3367 & 0.2280 & 0.3392 & 0.2282 & 0.3407 & \textbf{0.2217} & \textbf{0.3355} \\
\midrule
\multirow{4}{*}{Exchange} & 96 & 0.0610 & 0.1745 & 0.0598 & \textbf{0.1704} & \textbf{0.0596} & 0.1720 & 0.0601 & 0.1715 & 0.0607 & 0.1721 \\
& 192 & \textbf{0.1157} & \textbf{0.2455} & 0.1259 & 0.2525 & 0.1292 & 0.2555 & 0.1393 & 0.2667 & 0.1260 & 0.2524 \\
& 336 & \textbf{0.2024} & \textbf{0.3408} & 0.2609 & 0.3726 & 0.2535 & 0.3668 & 0.2640 & 0.3766 & 0.2663 & 0.3751 \\
& 720 & \textbf{0.3464} & \textbf{0.4646} & 0.8420 & 0.7169 & 0.7973 & 0.6985 & 0.8628 & 0.7295 & 0.8415 & 0.7170 \\
\midrule
\multirow{4}{*}{Electricity} & 96 & 13.2793 & 0.3959 & 104.0162 & 0.7849 & \textbf{12.9196} & \textbf{0.3914} & 12.9662 & 0.3947 & 194.9510 & 0.9955 \\
& 192 & 15.6661 & 0.4281 & 92.0825 & 0.7862 & \textbf{15.2447} & 0.4313 & 15.3682 & 0.4367 & 90.6173 & 0.7797 \\
& 336 & 22.3238 & 0.4890 & 56.5880 & 0.7061 & \textbf{17.9996} & \textbf{0.4462} & 18.5211 & 0.4703 & 38.7440 & 0.6173 \\
& 720 & 24.9721 & 0.4577 & 36.7740 & 0.6155 & \textbf{24.9073} & \textbf{0.4580} & 24.8309 & 0.4644 & 31.4215 & 0.5791 \\
\bottomrule
\end{tabular}%
}
\end{table*}

The comprehensive results in Table~\ref{sec:final_verdict} deliver the final verdict on our exploration of diagnostics-driven adaptive strategies. The data reveals two crucial insights. First, the naive \texttt{R$^2$-IN}, while not always the single best performer in every task, consistently avoids catastrophic failure and maintains a strong, stable performance, particularly on the most challenging \textit{Electricity} dataset.

Second, and most critically, the performance of the \texttt{Final\_A\_IN} model provides the most profound insight. This model was statically pre-configured with what our diagnostic framework deemed the "optimal" strategy for each dataset. As shown in the table, on the \textit{Electricity} dataset, where it was configured to use \texttt{RevIN}, it perfectly replicated \texttt{RevIN}'s catastrophic failure across all forecast horizons. This finding is definitive: the model's failure was not a failure of the diagnostic data, but a failure of the counter-intuitive heuristic built upon it. It correctly executed its flawed logic. The model's intelligence was rendered meaningless because the rule it was given—'when risk is high, use the most sensitive tool'—was fundamentally wrong for the job.

This elevates our "less is more" conclusion to a core principle: the pursuit of complex, adaptive normalization schemes for simple linear models is not only unnecessary but can be actively detrimental. The consistent, top-tier performance of the simple, non-adaptive \texttt{R$^2$-IN} across all benchmarks is not an anomaly; it is a clear signal that for time series forecasting with linear models, unwavering simplicity and robustness are the most direct path to state-of-the-art performance.

\subsection{Qualitative Analysis and Case Study}
\label{sec:qualitative}

To offer deeper insight beyond quantitative metrics, we present a qualitative case study on a representative sample from the \textit{Electricity} dataset where we simulate an extreme outlier in the lookback window.

\begin{figure}[t]
    \centering
    \includegraphics[width=\columnwidth]{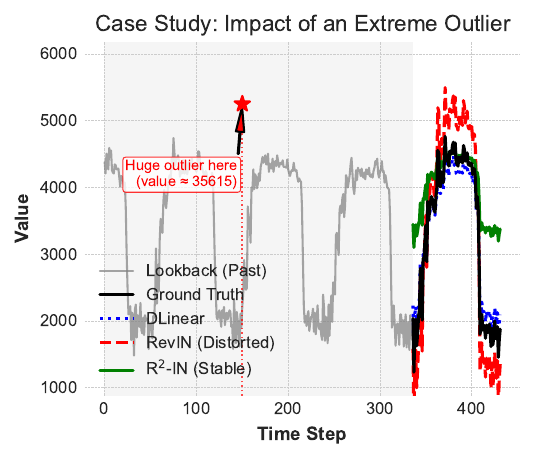} 
    \caption{A case study on a sample from the Electricity dataset. An extreme outlier (indicated by the red star) in the lookback window catastrophically skews RevIN's statistics, distorting its forecast. The robust \texttt{R$^2$-IN} ignores the outlier and produces a stable prediction.}
    \label{fig:case_study}
\end{figure}

Figure~\ref{fig:case_study} visualizes the resulting forecasts. The outcome is striking: the outlier contaminates the statistics of \texttt{RevIN}, causing its forecast to be severely shifted and distorted. In stark contrast, the robust \texttt{R$^2$-IN} effectively ignores the anomaly, producing a stable forecast correctly aligned with the ground truth. This case study vividly illustrates the practical consequences of our theoretical deconstruction, providing clear visual confirmation of RevIN's fragility and the critical importance of robustness against real-world data anomalies.

\section{Discussion}
\label{sec:discussion}

\noindent\textbf{Main Findings and Implications.}
The central takeaway from our work is that \textbf{simplicity trumps complexity} in instance normalization. The failure of our A-IN model—caused by a counter-intuitive heuristic that applied a sensitive method to high-risk data—and the victory of the simple R$^2$-IN, starkly demonstrate that a poorly designed 'smart' system is worse than a simple, robust one. Our results advocate for a paradigm shift towards \textbf{diagnostics-driven model selection} over designing ever-more-complex modules. The surprising effectiveness of R$^2$-IN serves as a testament to a robust-by-default heuristic: for linear models, ignoring extreme events is often more effective than sensitively (and potentially incorrectly) adapting to them.

\noindent\textbf{Limitations.}
The primary limitation is the failure of our adaptive model, A-IN, which stems from its simple yet flawed heuristic of mapping high structural change risk to the sensitive RevIN strategy. While its failure demonstrates the peril of this specific design, it does not preclude the success of more sophisticated, learnable adaptive mechanisms. Furthermore, our study was restricted to a DLinear backbone, and findings may not generalize to more complex architectures.

\noindent\textbf{Future Work.}
Future work should investigate three critical questions: Why is the naive R$^2$-IN so effective? How can we develop truly learnable adaptive normalization (e.g., using meta-learning) to replace brittle heuristics? And do our findings generalize to more complex architectures?

\section{Conclusion}
\label{sec:conclusion}

In this paper, we addressed the unstable performance of instance normalization methods in time series forecasting. We began with a counter-intuitive empirical finding: the standard RevIN is prone to catastrophic failure, while its theoretically superior robust alternative, R$^2$-IN, exhibits inconsistent performance. This led us to deconstruct the underlying theory, where we identified four fundamental contradictions inherent in this class of methods.

To navigate these contradictions, we proposed a novel diagnostics-driven paradigm. We introduced a framework to profile time series data and developed two new methods: a theoretically corrected R$^2$-IN+ and a data-adaptive A-IN. Our extensive experiments led to a surprising and humbling conclusion: our proposed "smarter" methods failed to deliver. The corrected R$^2$-IN+ offered no significant overall advantage, and the adaptive A-IN performed the worst of all, as its \textbf{simple, counter-intuitive heuristic proved actively detrimental on high-risk datasets}. Counter-intuitively, the \textbf{simplest, theoretically flawed, naive R$^2$-IN proved to be the most robust and effective method overall.} 

The core contribution of this work is thus a cautionary tale and a new perspective: we must move from a blind search for complexity towards a diagnostics-driven analysis that appreciates the surprising power of simple, robust baselines and acknowledges the perilous nature of naively designed adaptive rules.

\section{Practical Recommendations for Practitioners}
\label{sec:appendix_guide}

We distill our findings into a practical guide for practitioners, advocating for a brief, upfront diagnostic step. The overall effectiveness of \texttt{R$^2$-IN} makes it a strong default choice, but for optimal performance, a diagnostics-driven approach is paramount. The decision framework is summarized in Algorithm~\ref{alg:decision_framework_appendix}.

\begin{algorithm}[H]
\caption{A Practical Guide to Choosing an Instance Normalization Strategy. The algorithm provides a decision framework based on two lightweight diagnostic metrics computed on the dataset. First, it calculates the empirical k-factor ($k_{emp}$) to quantify the presence of extreme, non-Gaussian outliers, and the Change Point Risk (CPR) to measure structural instability. Based on these metrics, the rules are as follows: if $k_{emp}$ is very high (e.g., $>1000$), a robust method (\texttt{R$^2$-IN} or \texttt{R$^2$-IN+}) is strongly preferred. If CPR is high (e.g., $\ge 0.75$), the stable \texttt{R$^2$-IN} is a cautious choice. For relatively "well-behaved" data, both \texttt{RevIN} and \texttt{R$^2$-IN} are considered viable candidates for evaluation. Finally, if no diagnostics are performed, the algorithm defaults to \texttt{R$^2$-IN} as it was found to be the safest and best overall baseline.}
\label{alg:decision_framework_appendix}
\begin{algorithmic}[1]
\State \textbf{Input:} A time series dataset $\mathcal{D}$.
\State \textbf{Output:} A recommended normalization strategy.
\Procedure{SelectNormalization}{$\mathcal{D}$}
    \State Compute average empirical k-factor: $k_{emp} \gets \text{mean}\left(\frac{\text{std}(\mathbf{x})}{\text{MAD}(\mathbf{x})}\right)$
    \State Compute Change Point Risk: $CPR \gets \text{frequency of recent change points}$
    \Statex
    \If{$k_{emp} > 1000$} \Comment{Very high outliers}
        \State \Return \texttt{R$^2$-IN} or \texttt{R$^2$-IN+}
    \ElsIf{$CPR \geq 0.75$} \Comment{High structural instability}
        \State \Return \texttt{R$^2$-IN}
    \Else \Comment{Well-behaved data}
        \State \Return Evaluate both \texttt{RevIN} and \texttt{R$^2$-IN}
    \EndIf
    \Statex
    \If{\textnormal{no diagnostics available}} \Comment{Fallback}
        \State \Return \texttt{R$^2$-IN}
    \EndIf
\EndProcedure
\end{algorithmic}
\end{algorithm}

\bibliographystyle{ACM-Reference-Format}
\bibliography{sample-base}


\end{document}